\documentclass{article} % For LaTeX2e
\usepackage{colm2024_conference}

\colmfinalcopy

\usepackage{microtype}
\usepackage{hyperref}
\usepackage{url}
\usepackage{booktabs}
\usepackage{amsmath,amsthm}
\usepackage{amssymb}
\usepackage{graphicx}
\usepackage{textcomp}
\usepackage{tcolorbox}
\usepackage{soul}
\usepackage{makecell}
\usepackage[normalem]{ulem}
\usepackage{xspace}
\usepackage{arydshln}
\usepackage{xcolor}
\usepackage{arydshln}
\usepackage{setspace}
\usepackage{wrapfig}
\newcommand{\modelname}{\mbox{\textsc{dCluB}}\xspace}
\definecolor{mypink}{RGB}{253, 238, 238}
\definecolor{myblue}{RGB}{217, 231, 253}
\definecolor{mygray}{gray}{0.5}

\usepackage[shortlabels]{enumitem}

\title{
\includegraphics[scale=0.02]{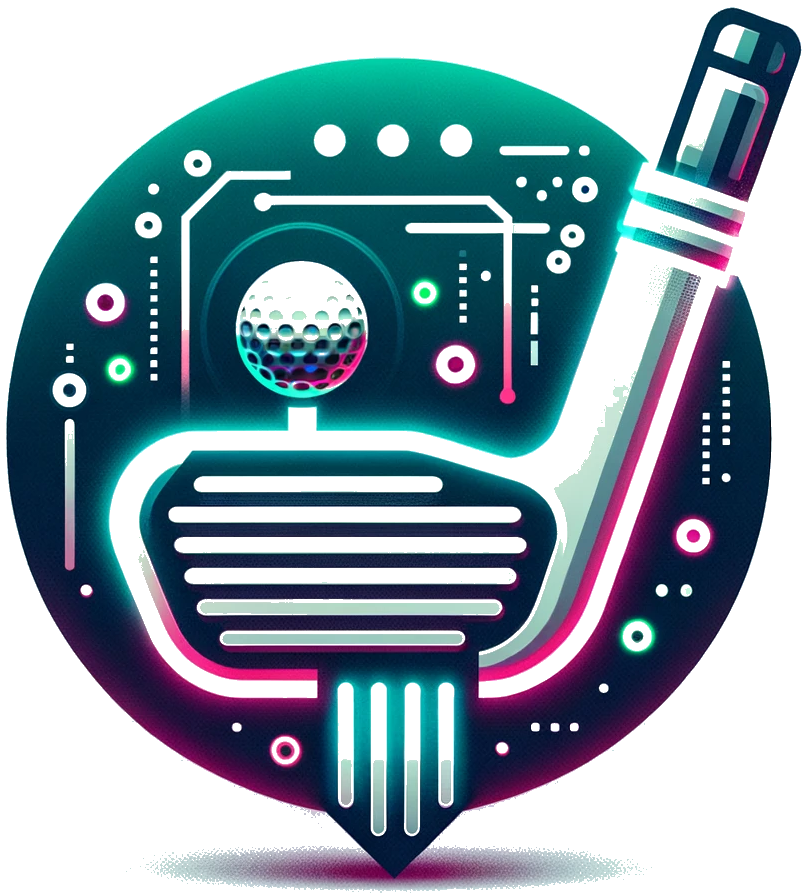}Dynamic Clue Bottlenecks:\\Towards Interpretable-by-Design Visual Question Answering}

\author{Xingyu Fu, Ben Zhou, Sihao Chen, Mark Yatskar, Dan Roth\\
  {University of Pennsylvania \;\;\;}\\
  \{xingyuf2, xyzhou, sihaoc, myatskar, danroth\}@seas.upenn.edu\\
  \color{purple}\url{https://zeyofu.github.io/dCluB}
}

\begin{document}

\maketitle

\begin{abstract}
Recent advances in multimodal large language models (LLMs) have shown extreme effectiveness in visual question answering (VQA). However, the design nature of these end-to-end models prevents them from being interpretable to humans, undermining trust and applicability in critical domains.
While post-hoc rationales offer certain insight into understanding model behavior, these explanations are not guaranteed to be faithful to the model.
In this paper, we address these shortcomings by introducing an  \textbf{interpretable by design} model that factors model decisions into intermediate human-legible explanations, and allows people to easily understand why a model fails or succeeds.
We propose the Dynamic Clue Bottleneck Model (\modelname \includegraphics[scale=0.01]{figures/dclub_icon.png}), a method that is designed one step towards an inherently interpretable VQA system. \modelname provides an explainable intermediate space before the VQA decision and is faithful from the beginning, while maintaining comparable performance to black-box systems.
Given a question and an image, \modelname first returns a set of \textbf{\textit{visual clues}}: natural language statements of visually salient evidence from the image, and then generates the output based solely on the visual clues. %, forming a dynamic information bottleneck.
% since visual clues are question and image specific.
% The core of our approach borrows the concept of information bottleneck, which limits the amount of information that can be explored during the reasoning and inference stage, posting a hard constraint on what the models can rely on and forcing the models to be interpretable and faithful.
% Crucially, relevant visual clues are question and image specific, and must be dynamically constructed to be effective.
% Evaluations show that zero-shot prompted \modelname could obtain around 95\% in average of the black-box VQA performance. for VQA v2 and GQA benchmarks.
To supervise and evaluate the generation of VQA explanations within \modelname, we collect a dataset of 1.7k questions with visual clues.
Evaluations show that our inherently interpretable system can improve 4.64\% over a comparable black-box system in reasoning focused questions while preserving 99.43\% of performance on VQA-v2. % and 95.24\% on GQA.
% Overall, our approach shows it is possible to design and supervise interpetable-by-design VQA systems that can achieve comparable performance to black-box systems. 

% Findings.
% 1. Interpretability is an additional requirement on the model, model will lose some part of performance because of this? (example: color example)

\end{abstract}

% \subsection{Citations within the text}
% When the authors or the publication are included in the sentence, the citation should not be in parenthesis using \verb|\citet{}| (as
% in ``See \citet{Vaswani+2017} for more information.''). 
% Otherwise, the citation should be in parenthesis using \verb|\citep{}| (as in ``Transformers are a key tool
% for developing language models~\citep{Vaswani+2017}.'').

\section{Introduction}
\begin{wrapfigure}{r}{0.5\textwidth}
  \begin{center}
  \vspace{-5em}
    \includegraphics[width=0.48\textwidth]{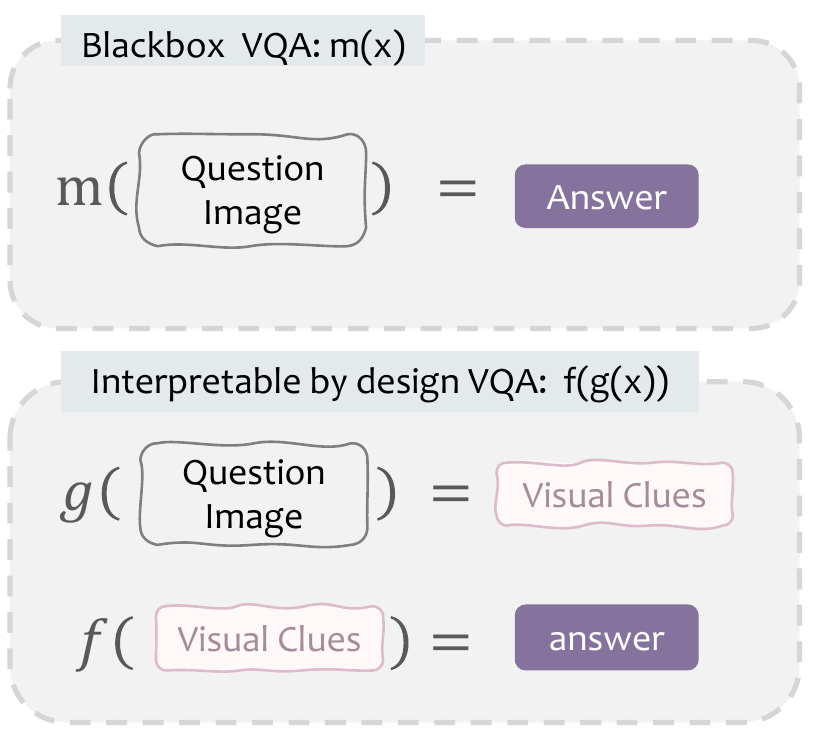}
    \label{fig:position}
  \end{center}
  \vspace{-1em}
  \caption{Design differences between de facto blackbox VQA methods (up), and our proposed \modelname method (bottom). Default models directly generate answers, while \modelname first provides visual clues in the image that could hint an answer, and then decides the answer based soly on the clues.}
  % \vspace{-2em}
\end{wrapfigure}

% interpretability research is useful for many downstream tasks: surface failures, guide interventions, and explain important model behaviors. 
Recent advances in multimodal large language models (LLMs) have achieved significant improvements in multiple vision language tasks, especially visual question answering (VQA)
% , where humans chat with a seemingly omnipotent and intelligent model
~\citep{liu2023improvedllava,OpenAI2023GPT4TR,liu2023llava,li2023blip,instructblip}.
However, these end-to-end models are not wholly trustworthy  because the computation processes are not interpretable, transparent or controllable, resulting in limited applicability to critical domains~\citep{rudin2019stop}.
Efforts to address this have largely focused on post-hoc explanations~\citep{okvqa,schwenk2022okvqa}. 
Recent studies indicate that post-hoc rationales by blackbox models seems plausible and might potentially reveal certain insight into their predictions, increasing trust in such models~\citep{selvaraju2016grad,park2018multimodal,okvqa}.
Problematically, post-hoc explanations are often systematically unfaithful to the answers, \eg models can have wrong rationals while correct answers~\citep{lyu-etal-2023-faithful} or misleading and unstable rationales with different inputs~\citep{turpin2024language}.

Models can also be designed to be inherently interpretable, helping to surface failures and to guide interventions. However, it is believed that such models will perform much worse than their black box alternatives~\citep{gunning2019darpa}. 
In this work, we try to answer the following question: Can we design an interpretable-by-design VQA system that provides faithful
% and human-legible explanations during the intermediate stage, 
explanations and performs as good as its blackbox counterpart? 
We provide evidence to answer ``yes'' to this question, and show how to construct a VQA system that is high-performance,  \textbf{inherently faithful} and \textbf{interpretable-by-design}.
To achieve this, we develop interpretable models where :
\begin{itemize}
  \item No secondary model is used to provide explanation.
  \item Model final prediction is entirely based on the explanations.
\end{itemize}
  % \item Model provides human-readable explanations.
  % \item Explanations are intermediate and not post-hoc.
  % \item Model final prediction is limited to base on explanations only.
  % \item No secondary model is used.
% with the purpose to answer the following questions:
% \begin{enumerate}[start=1,label={[\bfseries Q\arabic*]:}]
%   \item When does a model fail or succeed?
%   \item What factors determine the model prediction, and how?
%   \item How can we improve upon state-of-the-art (sota) models?
% \end{enumerate}

\begin{figure*}[t]
  \centering
  \includegraphics[width=.95\textwidth]{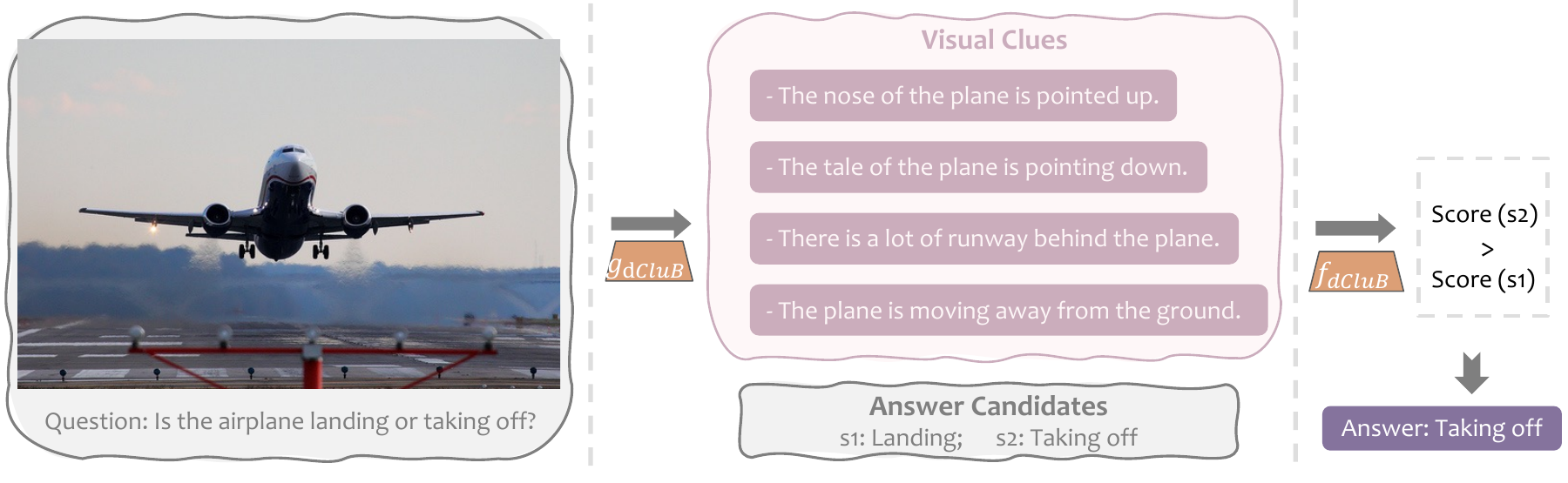}
  \caption{A detailed illustration of our \modelname system on an example VQA data, with explicit steps of visual clue generation using $g$ and natural language entailment scores from $f$ for final prediction. Answer candidates are pre-given in our setting, and we use top-k answers from the counterpart blackbox model as answer candidates in our experiments.}
  \label{fig:example}
  % \vspace{-em}
\end{figure*}

% As far as this paper is written, no inherently interpretable VQA models has been designed.
% The major challenge of such models is that they tend to dramatically under-perform their black-box counterparts, and that the human understandable intermediate structure for VQA is hard to specify or define and remains unclear.

We propose the \underline{D}ynamic \underline{Clu}e \underline{B}ottleneck Model (\modelname \includegraphics[scale=0.01]{figures/dclub_icon.png}) for \textbf{interpretable by design} visual question answering. 
Our system design takes inspiration from the concept bottleneck model ~\cite{pmlr-v119-koh20a}.
% , a new type of concept bottleneck model~\cite{pmlr-v119-koh20a}, appropriate for VQA. 
Unlike blackbox VQA systems, \modelname is interpretable because it factors prediction into two stages: a human readable information bottleneck and a predictor that only conditions on the bottleneck.
Abstractly, as shown in Figure \ref{fig:position}, our bottleneck model factors the full predictor as $y = f(g(x))$, where $g$ is the bottleneck predicting function, and $f$ forms the final prediction.
$g$ is allowed to be arbitrarily complex, adding expressivity, but $f$ must be easy for a person to understand, enforcing interpretability. 
Previous work manually designed static bottleneck $g$ or specified it with common sense from language models~\cite{yang2023language}, for image classification.

% VQA is uniquely challenging and subsumes classification: information relevant for answering a question is image dependant, the connection between information and answers may be under-specified, and require abstract reasoning.
% With a flexible design allowing too much information, these models will fall back to semantic biases.
% On the other hand, with a controlled design that allows for too little information, models cannot sufficiently and efficiently solve the task. 
% The core of our approach borrows the concept of information bottleneck, which limits the amount of information that can be explored during the reasoning and inference stage, posting a hard constraint on what the models can rely on and forcing the models to be interpretable and faithful. 

As illustrated in Figure \ref{fig:example}, we address the challenges of building bottlenecks for VQA by creating a system $g$ that tries to generate simple natural language arguments that may be useful for answering a question. 
Possible answer candidates are pre-given in our setting, and we use top-k answers from the same blackbox VQA model that we compare with in our experiments.
The final predictor, $f$, iterates over the generated visual clues with a simple NLI system evaluating how much visual clues support each answer proposal and returning the answer with highest support.

% We are predicting following f(g(x)). g is... Novelty: g is dynamic since question and image based, and supervised through our collection effort.

We also collect a dataset of 1.7k VQA instances requiring reasoning~\citep{selvaraju2020squinting, balanced_vqa_v2} to annotate with visual clues.
The dataset is used to fine tune our visual clue generator $g$, and as a reasoning focused test set.
In comparison to a BLIP-2~\citep{li2023blip} model fine tuned on equal data, \modelname improves by \textbf{4.64}\% on our reasoning focused test set, and maintains 99.43\% of performance on benchmark data from VQA v2~\citep{balanced_vqa_v2}, while offering better interpretability.
Additional experiments show that if we use a zero-shot MLLM such as LLaVA-v1.5~\citep{liu2023improvedllava} to replace our supervised bottleneck $g$, it would be 9.98\% worse, proving efficacy of our fine tuned visual clue generator $g$.
Our results demonstrate a promising direction toward building faithful and interpretable by design multi-modal systems that perform as well as their black-box counterparts. 

To summarize, the contribution of our work are as follows: 1) We create an interpretable by design VQA system by formulating an effective bottleneck structure 2) Our system \modelname achieves comparable performance as its blackbox counterparts while providing faithful explanations; 3) We collected a 1.7k dataset for training and evaluation with VQA explanations.

\section{Related Works}
\subsection{VQA Interpretability}
Some earlier attempts to rationalize VQA decisions~\citet{xiong2016dynamic,shih2016look,das2017human} try to answer the question ``Where should we look at the image to answer the question?'' through attention maps. However, it is unclear how focusing on certain parts of the image help answer the question. 
Several works in this direction have achieved higher interpretability by generating visual attention scores as intermediate steps of black box models~\citet{xiong2016dynamic,shih2016look,das2017human,hendricks2018grounding,anderson2018bottom}.
However, attention scores are not interpretable to humans since they cannot clearly state how the attended area connects to the final answer.
Some recent VQA datasets are designed to encourage interpretability \citep{Fu2022TheresAT,schwenk2022okvqa}.
More recent works generate post-hoc natural language explanations~\citet{hendricks2016generating,dua2021beyond,schwenk2022okvqa,fu2023generate}, so as to reason between vision and language inputs at the same time.
However, the post-hoc explanations are not proven to be the exact reasons for the model predictions and thus are difficult to use to guide interventions or bring improvements. 

\subsection{Textual Interpretability}
With the rapid development of LLMs, the community has spent major effort on language model reasoning \citep{cot,creswell2023selectioninference,yao2023react,hong2023faithful}.
Some papers propose a decomposition process with LMs~\citet{Zhou2022LearningTD, khattab2022demonstrate}, or use post-hoc explanations from LLMs as a training or inference signal~\citet{Feng2022GenericTR}.
These methods have shown large improvement and huge potential, but are specific to different problems and cannot be directly applied to the visual question answering area.
It is worth mentioning that~\citet{pmlr-v119-koh20a,yang2023language} made attempts to design high-performance interpretable-by-design image classifier systems with the help of LLMs~\cite{gpt3}. However, this method is limited to image classification and cannot be easily transformed to the visual question answering problems.
\section{Dynamic Clue Bottlenecks VQA}
\label{sec:method}
As illustrated in Figure \ref{fig:position}, blackbox VQA methods in general learn a function $m$ for the answer prediction $y = m(x)$ where x denotes the question and image input. In contrast, our \modelname composes two functions $f$ and $g$ and predicts following $y = f(g(x))$, where $g(x)$ is our bottleneck model induced by visual clues, and $f$ is a natural language inference (NLI) function. 
To better illustrate our overall method, we first define the aforementioned term as following.

\begin{wrapfigure}{r}{0.5\textwidth}
  \begin{center}
  \vspace{-2em}
    \includegraphics[width=0.48\textwidth]{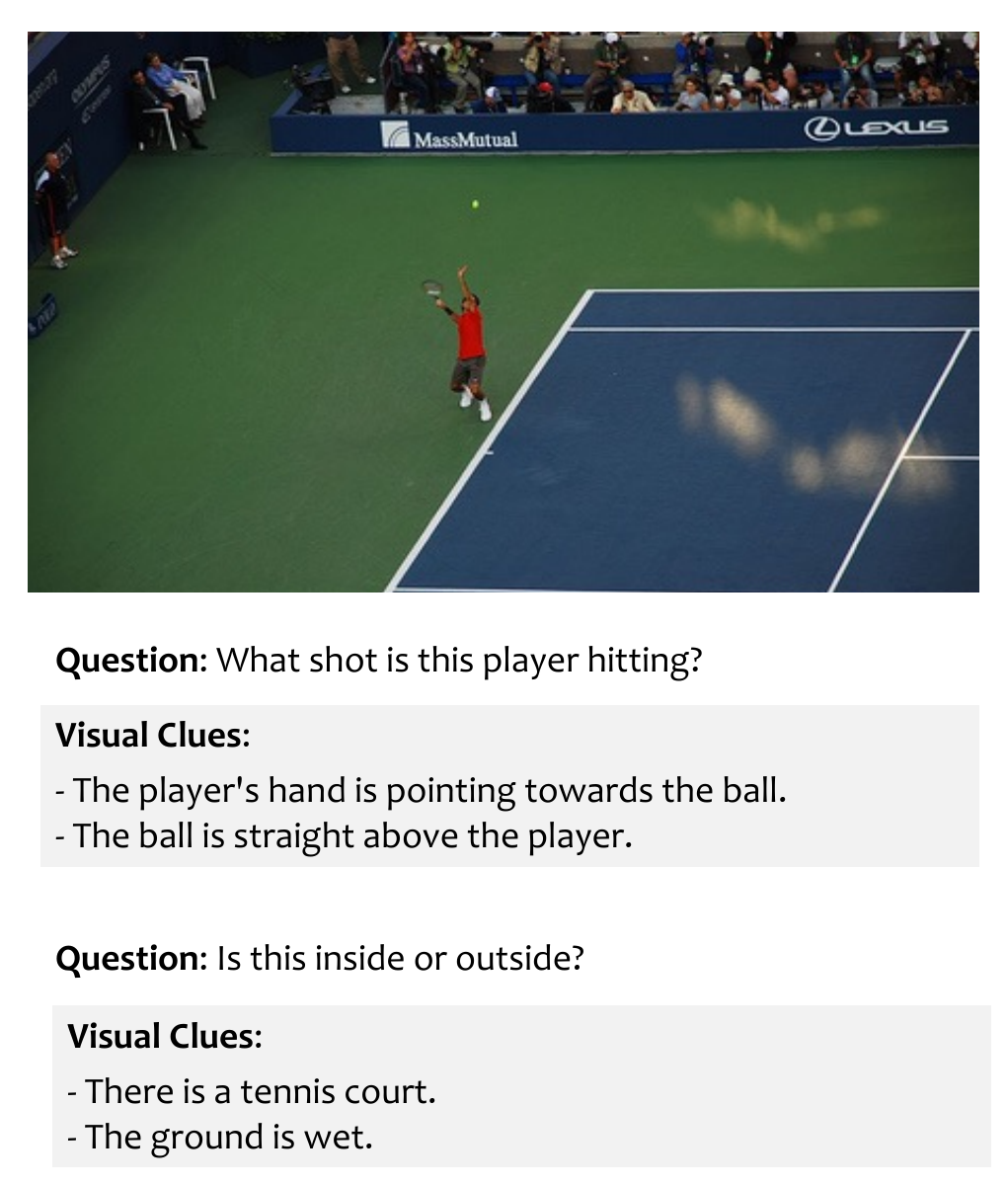}
    \label{fig:clue_example}
  \end{center}
  \caption{Examples showing the dynamic nature of \modelname's visual clues: for the same image, different questions should have different visual clues.}
 % \vspace{-1em}
\end{wrapfigure}

\textbf{Visual Clues.} As illustrated in Figure~\ref{fig:collected_data}, we define \textit{visual clue} as natural language descriptions that help to answer the question while entirely grounded inside the corresponding image. This is the intended output of $g(x)$.
For clarification, our visual clues are dynamic and should be different for cases with same question and different images, or with same image and different questions. Example can be found in Figure~\ref{fig:clue_example}.

Given a visual question answering pair $(v, q, a)$ where $v$ denotes the image, $q$ denotes the question, and $a$ represents the answer respectively, we train a separate multimodal model that generates visual clues grounded in $v, q$. 
Given a set of possible answer proposals $\hat{A} =\{\hat{a_1}, \hat{a_2}, ...\}$, we deduce the final answer based on the visual clues.

\subsection{Visual Clue Generator $g(x)$}
This step aims to develop a visual-language model to reason over the image and question and generate visual clues that help lead to an answer. Specifically we define 
$$visual\_clues(v, q)= g(v,q)$$ 
We implement this function with a fine-tuned BLIP-2 model~\citet{li2023blip} following similar settings as in \citet{instructblip}. Specifically, we provide the Q-former module with the question to help it better extract image features and include the question again in input to the frozen LLM inside BLIP-2.
% , as illustrated in Figure \ref{fig:blip}. 
The frozen LLM input includes image features and a prompt as ``\texttt{Question: \{\textit{$q$}\}. Visual Clues: {}}'' during both training and inference. 

Note that there are some differences between our visual clue generator and the original BLIP-2 model. When the original BLIP-2 model was trained for image captioning, each image with $N$ gold captions will compose $N$ input data with one single caption in each training input. In our case, we find that the generated clues trained this way always share high similarity and have same starting tokens. Instead, for a data with $N$ gold visual clues, we follow previous works \citet{klein-etal-2022-qasem,chen-etal-2023-propsegment} and compose $N$ training inputs, each containing one unique permutation of concatenated visual clues.
The BLIP-2 clue generation model is trained with standard language modeling loss to directly generate all the visual clues at the same time given an image and question. This strategy is used to encourage the diversity and comprehensiveness of generated clues.

\subsection{Final Predictor $f$}
With visual clues generated, we concatenate all of them together and ask a natural language inference (NLI) model to rate the likelihood of the candidate answers.
Notice that in our setting, the answer proposals $\hat{A} =\{\hat{a_1}, \hat{a_2}, ...\}$ are pre-given. In our experiments, the candidate answers are top-k answers from the blackbox counterparts (\eg, we finetune a base model with visual clues training data and get $g(x)$, then we finetune the same base model with VQA pairs of the same set of training data as the blackbox conterpart).
We first turn the proposals and the question into statements. For instance, for the question ``\texttt{Was the photo taken during the day?}'' with answer proposals being ``\texttt{Yes; No}'', we get the candidate answer statements as ``\texttt{The photo was taken during the day; The photo was not taken during the day}''.
They we deploy a NLI model to determine the score for each candidate answer statement:
\[
    score(\hat{a_i})= entail(visual\_clues(v, q), statement(q, \hat{a_i})) 
\]
where $\hat{a_i}$ is the $i^{th}$ answer proposal. $score(\hat{a_i})$ rates how likely the statement can be true given the visual clues on a scale of 1 to 9, with 1 being not likely and 9 meaning almost surely true. 
The final answer prediction $\hat{y}$ is simply decided by selecting the answer proposal with highest score:
\[
\hat{y} = \underset{\hat{a_i}}{\text{argmax}} \, score(\hat{a_i})
\]
The NLI is conducted using a LLaMa~\citep{touvron2023llama} model~\footnote{We use the LLaMa-2-70B-instruct model from \url{https://huggingface.co/upstage/Llama-2-70b-instruct}} under few-shot prompting, and the prompt we use can be found in Figure~\ref{fig:prompt_entail}.

\section{Visual Clues Dataset Collection}
\label{sec:data}
\begin{figure*}[t]
  \centering
  \includegraphics[width=1\textwidth]{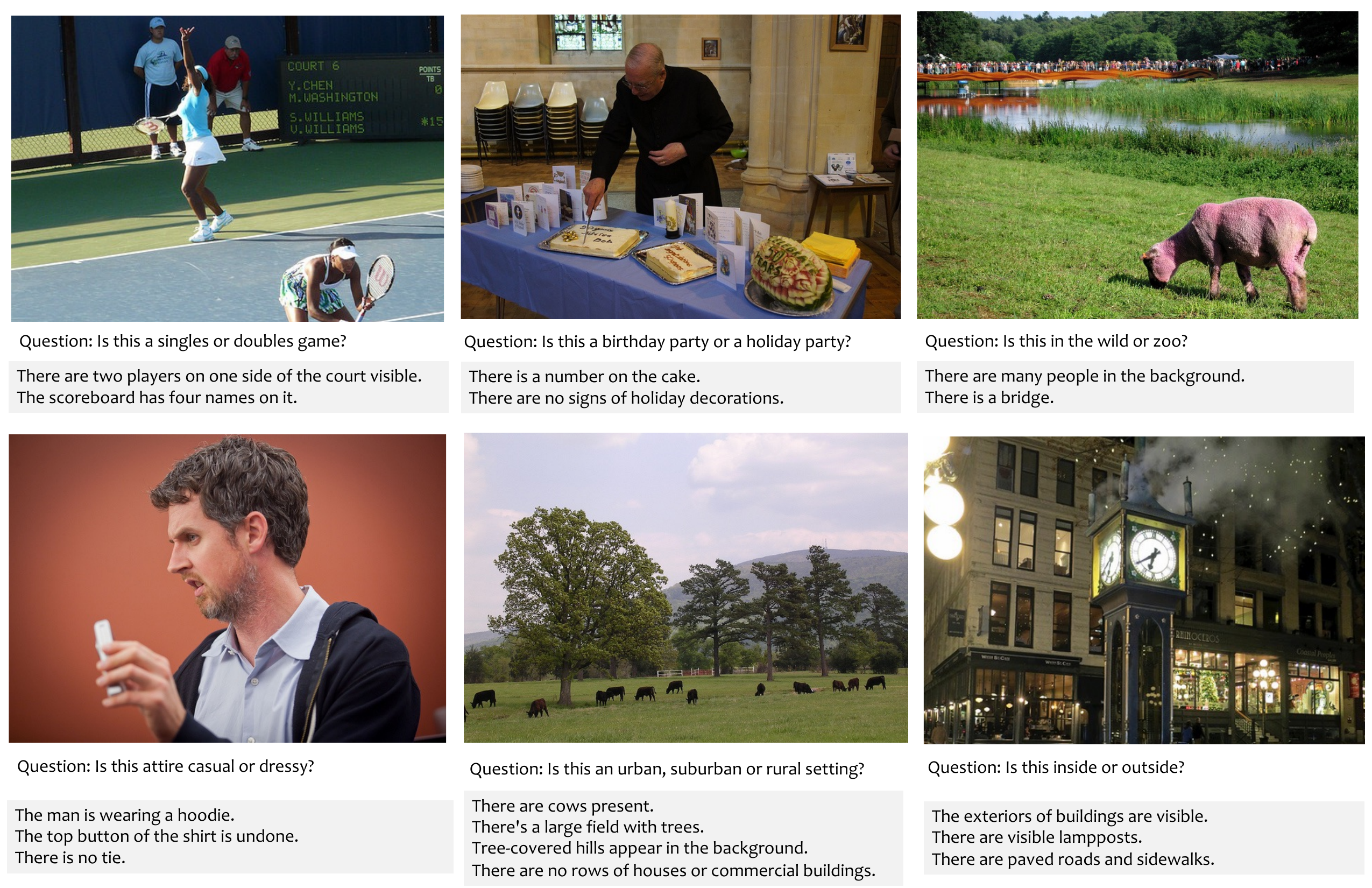}
  \caption{Collected visual clues examples in our training data.}
  \label{fig:collected_data}
  % \vspace{-em}
\end{figure*}
Since there is no existing visual clue data ready for our proposed \modelname system, we use Amazon Turk to collect 1.7K high-quality data for learning, evaluation, and analysis purposes, while focusing on questions that require reasoning besides simple recognition or perception following~\citet{selvaraju2020squinting}.
Given a question, image, and answer data pair, we ask the annotators to provide explicit visual clues that are entirely grounded in the image, and inferences that connect the visual clues to the answer. 
We also have additional filtering step of removing ambiguous questions by asking ``\texttt{if there is significant direct evidence in the image that supports a different answer}'' and deleting the questions with ``yes'' answers.

As illustrated in Figure~\ref{fig:collected_data}, the turkers always give us two to four visual clues for each VQA data pair. These clues are natural language bottlenecks, corresponding to salient visual evidence in an image that supports the answer to a question, but without directly telling the answers.
Dataset statistics can be found in Table \ref{tab:stat}. We try to have a more balanced set by maintaining a similar boolean question ratio (around 40\%) as in the VQA V2 dataset.
The detailed annotation guidelines and an example data entry are shown in Appendix \ref{sec:app:data}.
% \subsection{Dataset Quality}
% \label{subsec:dataset_quality}
% We test the annotated data quality using the same setting as in our experiments in Section \ref{sec:exp}, while replacing the generated visual clues with gold visual clues and additionally including gold answer in the answer candidates for fair comparison. Results in Table \ref{tab:anno_exp} show that in above 95\% of the time \modelname can select the correct answer with correct set of clues.

\begin{table}[h]
% \small
\centering
\begin{tabular}{c|cccc}
\toprule
Dataset & Train & Dev & Test & All\\ 
\midrule
\modelname & 1,143 & 302 & 291 & 1,736 \\
\bottomrule
\end{tabular}
\vspace{-0.5em}
\caption{Dataset statistics for the visual clues collected in \modelname. Details are in Section \ref{sec:data}.}
\label{tab:stat}
% \vspace{1em}
% % \end{table}
% % \begin{table}[h]
% \centering
% \begin{tabular}{lcc}
% \toprule
% Dataset & Dev Acc & Test Acc\\
% \midrule
% \modelname & 95.03 & 96.56\\
% \bottomrule
% \end{tabular}
% \caption{VQA performance using gold visual clues. Details can be found in Section \ref{subsec:dataset_quality}.}
% \label{tab:anno_exp}
\end{table}
% \begin{table}[h]
% \centering
% \begin{tabular}{lcc}
% \toprule
% Data & Dev Acc. & Test Acc. \\
% % \cmidrule(lr){1-1}\cmidrule(lr){2-2}\cmidrule(lr){3-3}
% \midrule
% \midrule
% \multicolumn{3}{c}{GPT-3.5}\\
% \midrule
% question only & 21.74 & 20.92\\
% visual clues & 62.69 & 64.02\\
% visual clues + inferences & 79.69 & 78.05\\
% \midrule
% \midrule
% \multicolumn{3}{c}{GPT-4}\\
% \midrule
% question only & 48.12 & 50.17\\
% visual clues & 82.75 & 82.82\\
% visual clues + inferences & 93.78 & 91.07\\
% \bottomrule
% \end{tabular}
% \caption{An analysis on our annotated \modelname data quality by using the gold visual clues and inferences to replace the image, and directly query GPT model for answer prediction.}
% \label{tab:gold}
% \end{table}
% In our second setting, we use Large Language models to see if they can select the correct answer given the gold visual clues only, without the image. 
% We concatenate the gold clues, and gold inferences together, as shown in Figure \ref{fig:prompt_vqa}. Note that we use 4-shot in-context learning examples retrieved from the training set according to question similarity by sentence bert. While the GPT-3.5 performance are not high enough, since the scores are approaching 100\% for GPT-4, we believe the annotated data, especially the visual clues, are high quality.

\begin{figure*}[ht!]
  \centering
  \includegraphics[width=1\textwidth]{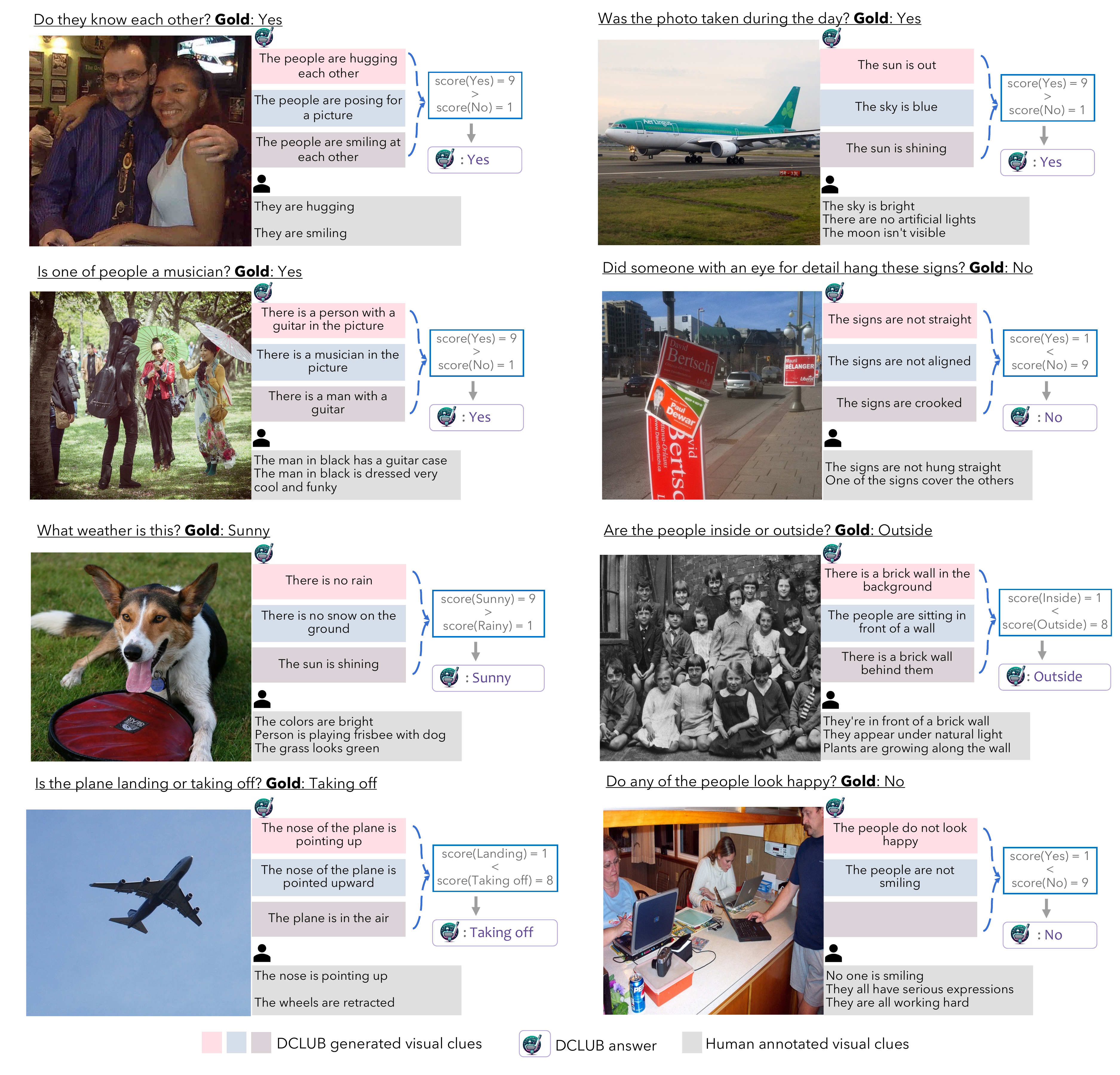}
  \caption{Qualitative examples outputs using \modelname \includegraphics[scale=0.01]{figures/dclub_icon.png}. Human annotated visual clues are in the grey boxes under the human icon, and \modelname visual clues are in colored boxes.}
  \label{fig:generated_clues_compare}
  % \vspace{-em}
\end{figure*}
\section{Experiments}
\label{sec:exp}
We evaluate on three datasets: our annotated \modelname dataset, the VQA v2~\citet{balanced_vqa_v2} benchmark, and GQA benchmark~\citet{hudson2019gqa}. Since our system depends on LLM model for entailment and is therefore limited by the LLM query speed, we randomly select a subset of size 300 from VQA v2 and from GQA for computation efficiency concerns.
% This section includes VQA end-task results, answer proposal results, and visual clue generation results.
In this section, we first describe the baseline models and experimental setup (\S\ref{sec:exp_setup}).
Then we present evaluation of both blackbox models and \modelname (\S\ref{sec:exp_results}). 
We demonstrate that as an interpretable-by-design VQA method, \modelname can achieve black-box model-level performances.
Finally, we provide detailed analysis on when and how \modelname succeeds or fails through intermediate outputs(\S\ref{sec:exp_analysis}).

\subsection{Blackbox Counterpart Baselines}
\label{sec:exp_setup}
We compare \modelname with its blackbox counterparts. Specifically, we fine-tune visual clue generator on a base pre-trained BLIP-2~\citet{li2023blip}, and it has no acceses to the gold VQA answer at any time. In comparison, the blackbox counterpart is the same base pre-trained BLIP-2 model fine-tuned on the same set of data, but with image-question-answer paired supervision only.
We deploy the pre-trained BLIP-2 checkpoint ``pretrain\_flant5xl'' in our experiments.

\textbf{Implementation Details}
We implement our visual clue generator based on the BLIP-2 model in LAVIS library~\citet{li2023blip} and follow the same fine-tuning settings as in the original paper. In our experiments, we adopt the image encoder ViT-L/14 from CLIP~\citet{radford2021learning}) and frozen LLM: FlanT5-XL (3B), by using the pre-trained checkpoints released from the original BLIP-2 paper. 
Training details including hyperparameters can be found in Appendix \ref{sec:app:hyper}. All models are trained utilizing NVIDIA RTX A6000 (48G) GPUs. The clue generation training can be completed within two days on average with single GPU resource.
Since the training data size is only about 1.1k, we augmented our training set with examples from existing dataset~\citet{selvaraju2020squinting} where each image-question pair are equipped with sub-questions and sub-answers. We turn the sub-questions and sub-answers into statements, and serve as low-quality visual clues for the original data.
We fine-tune our visual clue generation model following a two-stage training strategy: first on the large weak supervision set, then on \modelname.
\subsection{Main Results}
\label{sec:exp_results}
\begin{table*}[h]
\centering
\begin{tabular}{l|c|c|c}
\toprule
Dataset/Model & Blackbox & \modelname & Percentage\\
\midrule
Our & 71.48 & 74.8 & \textbf{104.64}\\
VQA-v2 & 58.11 & 57.78 & 99.43 \\
GQA & 42.00 & 40.00 & 95.24\\
\midrule
\bottomrule
\end{tabular}
\caption{VQA performance comparisons between blackbox (BB) baseline model: BLIP-2, and \modelname, which uses BLIP-2 fine-tuned visual clue generator. 
% Note that BLIP-2 is finetuned for the same amount of VQA question-answer pairs from our training set. 
Blackbox model performance coverage percentage (our / blackbox) is calculated on the right column, showing that \modelname can achieve comparison results to blackbox models consistently.}
\label{tab:main_blip}
\end{table*}

We report the end task performance calculated following the standard VQA evaluation metric in~\citet{antol2015vqa}, and show the performance comparisons between our interpretable vqa model and their blackbox counterparts. 
In Table \ref{tab:main_blip}, we demonstrate that \modelname achieves a comparable result with its blackbox counterpart on both of VQA v2 and GQA: covering 99.43\% and 95.24\% of the blackbox model performances corresponding, and even reaches a higher accuracy score on our collected testing set, achieving 104.64\% of the blackbox model performance.
\begin{wrapfigure}{r}{0.5\textwidth}
  \begin{center}
  \vspace{-1em}
    \includegraphics[width=0.48\textwidth]{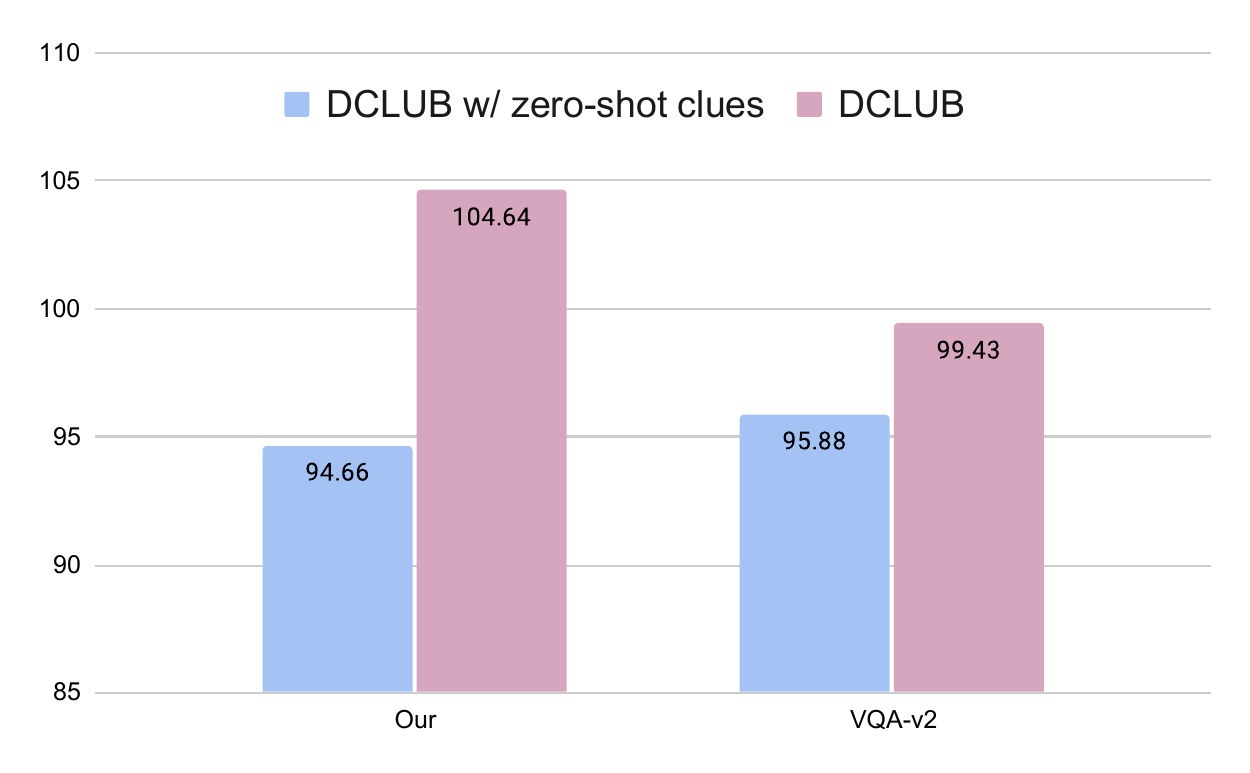}
    \label{fig:additional_llava}
  \end{center}
  \vspace{-2em}
  \caption{Comparison between the performance percentage coverage (\modelname / blackbox performance) using fine-tuned visual clue generator \vs zero-shot prompted visual clues. Compared with the performance coverages in Table~\ref{tab:main_blip}, we find that training a visual clue generator can help achieves better VQA results.}
  \vspace{-1em}
\end{wrapfigure}
Since \modelname has no access to the gold VQA answers of the training data during fine-tuning, we believe the percentage comparison illustrates the efficacy of visual clues in helping answer the VQA questions.

\begin{figure*}[h]
  \centering
  \includegraphics[width=1\textwidth]{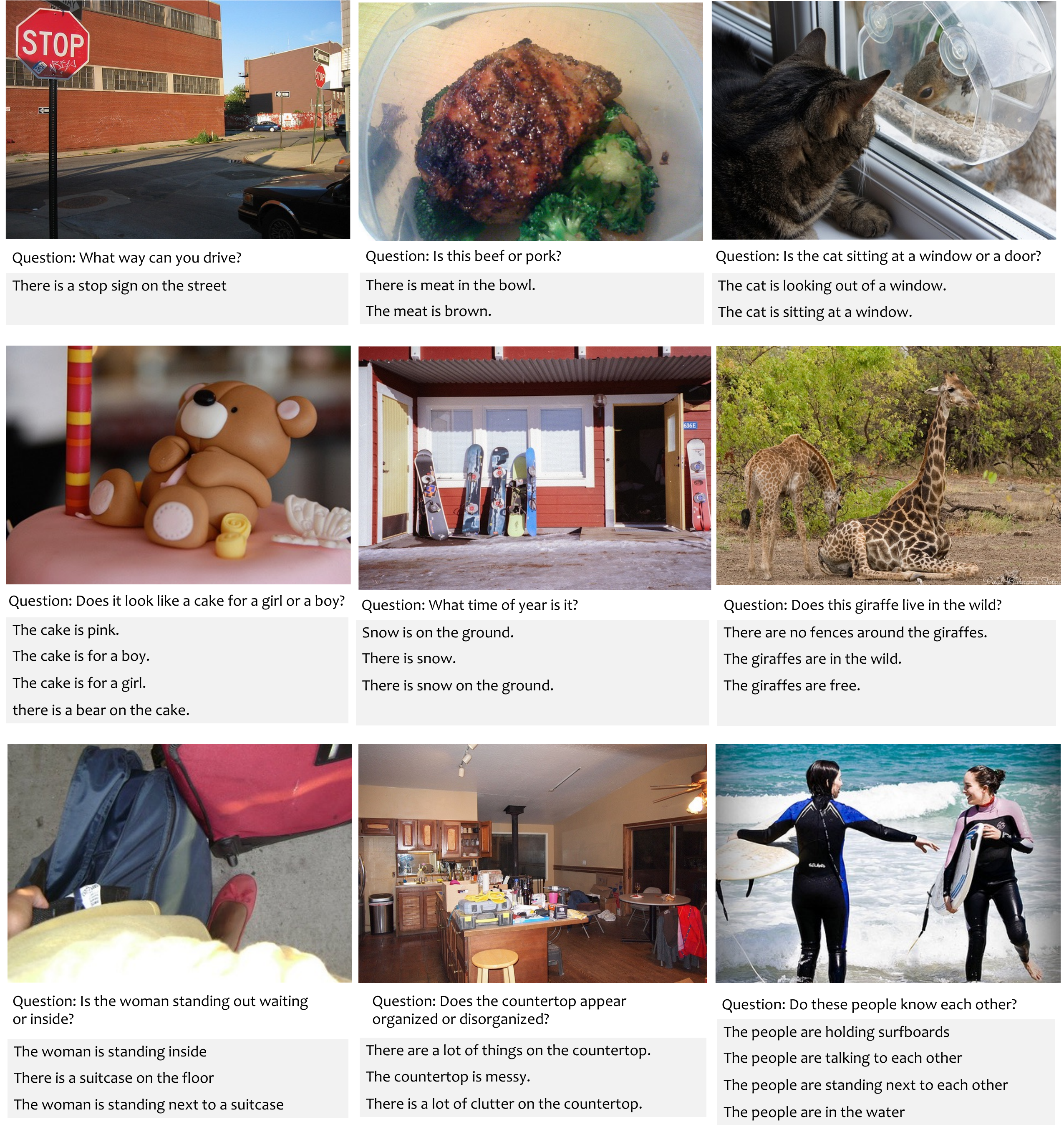}
  \caption{Examples of generated visual clues (in grey boxes) using \modelname.}
  \label{fig:generated_clues}
  \vspace{-1.2em}
\end{figure*}

\textbf{Demonstrating Generated Visual Clues}
We demonstrate a random select of generated clues examples from the \modelname data dev set as in Figure \ref{fig:generated_clues}.
From our observation, the overall clue quality is good, but sometimes the model
% struggles to avoid 
tends to directly include the answer to the question in the visual clues. 

\textbf{Additional Experiments with Zero-shot MLLM Explainer:}
An additional experiment is included in Figure~\ref{fig:additional_llava}, where we test the efficacy of the two-step pipeline versus blackbox end-to-end pipeline with the widely used open source MLLM model: LLaVa-v1.5-13b~\citep{liu2023improvedllava}.
Specifically, we compare LLaVa's blackbox VQA performance with that of interpretable-by-design VQA using LLaVa prompted visual clues following our pipeline.
The results show that using zero-shot LLaVa visual clues could obtain about 95\% LLaVa's blackbox performance on all three datasets, showing efficacy of the fine-tuned visual clue generator in \modelname.

\subsection{When does \modelname Succeed or Fail?}
\label{sec:exp_analysis}
% \begin{figure*}[h]
%   \centering
%   \includegraphics[width=1\textwidth]{figures/generated_clues_blip2_better.pdf}
%   \caption{Examples of cases where \modelname performs better than its blackbox counterpart. Examples are from the testing set.}
%   \label{fig:generated_clue_blip2_better}
%   % \vspace{-em}
% \end{figure*}
% \label{sec:exp_ans}

% \begin{figure*}[ht!]
%   \centering
%   \includegraphics[width=1\textwidth]{figures/generated_clues_blip2_worse.pdf}
%   \caption{Examples of cases where \modelname performs worse than its blackbox counterpart. Examples are from the testing set.}
%   \label{fig:generated_clue_blip2_worse}
%   % \vspace{-em}
% \end{figure*}

As an interpretable-by-design methods, \modelname provides intermediate human-legible explanations and can help humans understand why the model succeeds or fails.
From our observations, for cases where \modelname gives correct predictions, it succeeds mainly because it generates correct explanations (visual clues) supporting its prediction, as illustrated in Figure~\ref{fig:generated_clues_compare}.
However, it is unclear what happens when \modelname gives wrong predictions.

\begin{figure*}[t]
  \centering
  \includegraphics[width=1\textwidth]{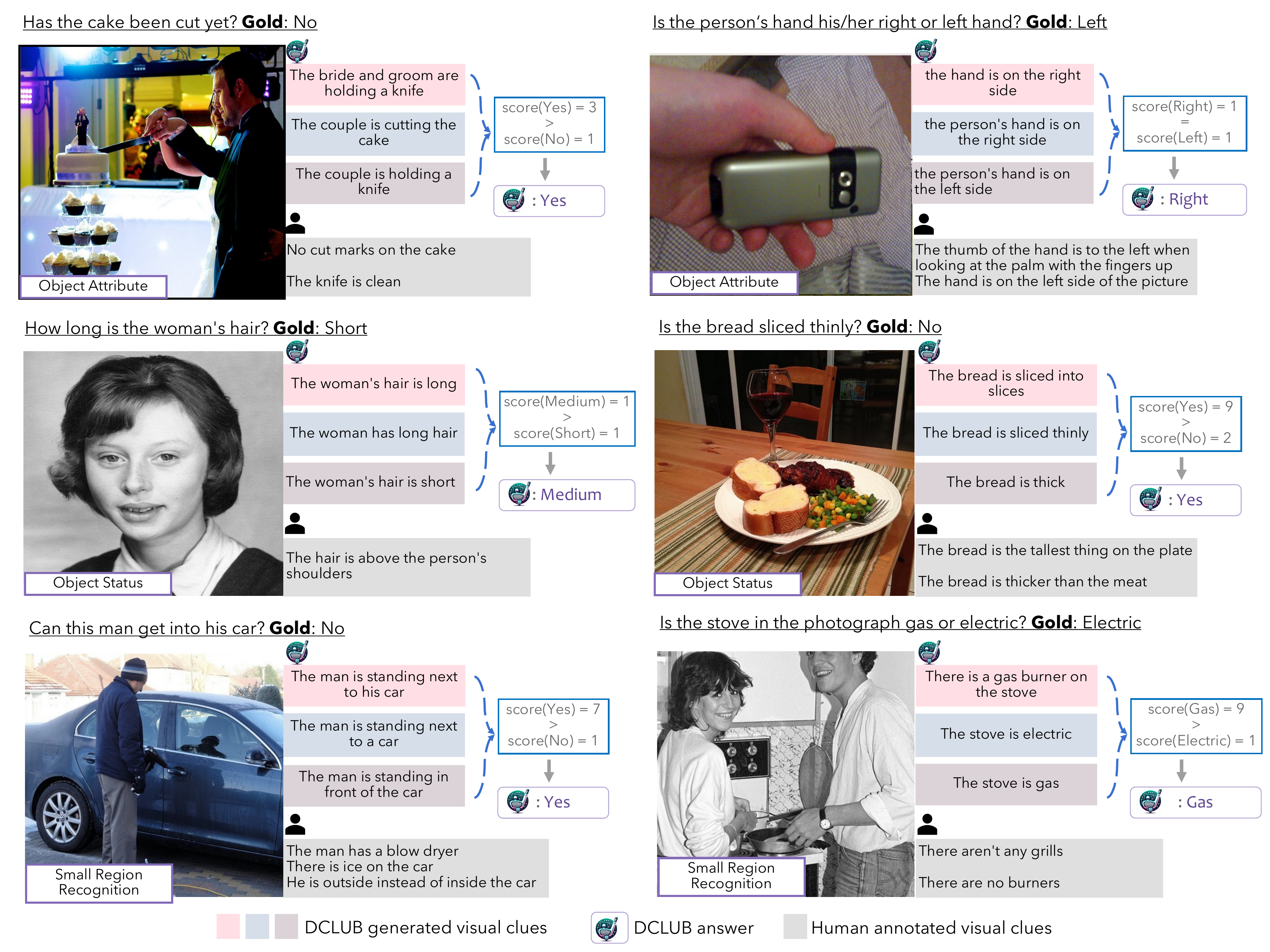}
  \caption{Examples of some common errors of \modelname: object fine-grained attributes, object status recognition, and small region recognition. Examples are from the validation set.}
  \label{fig:generated_clue_blip2_better}
  % \vspace{-em}
\end{figure*}

To answer this quesiton, we manually go through the error cases on the dev set of our collected data, and categorize the errors into three common reasons as illustrated in Figure~\ref{fig:generated_clue_blip2_better}-- 1. object attributes: as in the top row examples where the questions ask about a specific object attribute, \modelname fails to find the explanations for these fine-grained attributes; 2. object status: as in the second row examples in the figure, \modelname sometimes can hardly tell the status of an object \eg short or long, and thick or thin, and produces self-conflicting visual clues; and 3. small region recognition: as in the last row examples in the figure, the model only needs to focus on a relatively small region in the whole image to answer the question, \eg the blow dryer region and the stove burner region. However, \modelname tends to either neglect the small region, as in the blow dryer example, or directly tell the VQA answer, as in the stove example, resulting in bad visual clues and thus incorrect answers.

\label{sec:exp_ans}

\section{Conclusion}
In conclusion, we have presented \modelname, an inherently interpretable VQA system that breaks down the VQA process into visual clues, and entailment.
Our approach enables a deeper understanding of how the model arrives at a final answer while maintaining competitive performance compared with black-box VQA systems.
% To facilitate the development and evaluation of our approach, we have collected a high-quality 1.7k dataset using crowdsourcing, focusing on questions that require deeper reasoning abilities.
% Our results demonstrate the effectiveness of our approach in providing interpretable, step-by-step explanations that can be trusted by humans.
% This work represents a significant advancement in the field of interpretability in visual question answering, as it addresses the pressing need for more transparent, trustworthy, and understandable AI systems.
By making AI more accessible and comprehensible to users, we ultimately enhance the potential for beneficial applications across a variety of real-world domains, such as healthcare, finance, and more.

% In conclusion, this paper presents the \modelname Model, a novel concept bottleneck model tailored for visual question answering (VQA) tasks that addresses both the need for interpretability and competitive performance. By introducing a two-stage procedure consisting of a human-readable information bottleneck and a predictor dependent on the bottleneck, our approach enhances the inherent interpretability of multi-modal models. To overcome the various challenges posed by VQA, we employ a system for generating simple natural language arguments supporting the proposed answers. Furthermore, we contribute to the research community by creating a dataset of 1.7k VQA instances requiring reasoning, annotated with visual clues. Our experiments demonstrate that the proposed model not only retains an impressive 87.4\% performance on benchmark data from VQA v2 but also surpasses a BLIP-2 model fine-tuned on equal data by 1.5\% on our reasoning-focused subset. These findings pave the way for future research on inherently interpretable multi-modal systems that perform on par with their black-box counterparts while providing clearer explanation for their predictions, ultimately fostering enhanced trust in AI models.

\paragraph{Limitations}
% \fxy{depending LLM slow}.
% Our approach depends on extremely large OpenAI models, limiting how broadly such models could be adopted given the number of API calls required to create out bottleneck models.
% This may be cost prohibitive for some, and bears a significant environmental burden.
% Cost limitations also prevented us from validating many natural choices for prompting, and different model sizes.
% The VQA resources we evaluate on are limited to English. 
The pretrained model in our approach may be able to extend beyond English given their pretraining data, we did not investigate such directions because our focus was primarily developing interpretability methods. 
Our main evaluation is end performance, and we assumed the bottlenecks we generate are fluent, without deep evaluation.
Finally, our approach uses large language models for NLI, leaving the possibility that one component failure on new data negatively impacts the performance of the rest.

% Our randomly sampled VQA v2 subset size is relatively small compared to the whole val set size because of the economical and time concerns using GPT models.
% Also, we did not experiment with GPT-4 for budget concerns. 
% For the clue generation model, we did not experiment with the larger LLM: FlanT5 XXL which has better performance because of GPU computation power limitations.

\section*{Ethics Statement}

As introduced in Section \ref{sec:data}, we annotated the data using crowd-workers through Amazon Mechanical Turk.
They are voluntary participants who were aware of any risks of harm associated with their participation.
We require the workers to be located in either Australia, Canada, Great Britain or the United States such that they are English speakers. We also require the workers to have HIT Approval Rate (\%) for all Requesters' HITs greater than or equal to 95\%. 
All crowd-workers were compensated by a fair wage determined by estimating the average completing time of each annotation task.
Each worker earn \$7 per 10 queries and each query should take less than 2 minutes to annotate. 
%Example screenshots of the NYT data and our annotation interface can be found in Appendix \ref{sec:appendix-mturk}.

Finally, our methods rely on pretrained models that took may have had many negative environmental impacts during their training. 
During development of our models, we took care to avoid replicating such harms by needlessly retraining or finetuning models already available, although our continued dependence on large models may encourage new training runs of even larger models.

% \section*{Acknowledgements}


\bibliography{colm2024_conference}
\bibliographystyle{colm2024_conference}

\appendix

\section{Appendix}

% \subsection{Generated Visual Clues}
% We include examples of generated visual clues by \modelname as in Figure~\ref{fig:generated_clues}.
% \begin{figure*}[h]
%   \centering
%   \includegraphics[width=1\textwidth]{figures/generated_clues.pdf}
%   \caption{Examples of generated visual clues (in grey boxes) using \modelname.}
%   \label{fig:generated_clues}
%   \vspace{-1.2em}
% \end{figure*}

\subsection{Prompts used in \modelname}
\begin{figure}[h]
  \centering
  \includegraphics[width=0.5\textwidth]{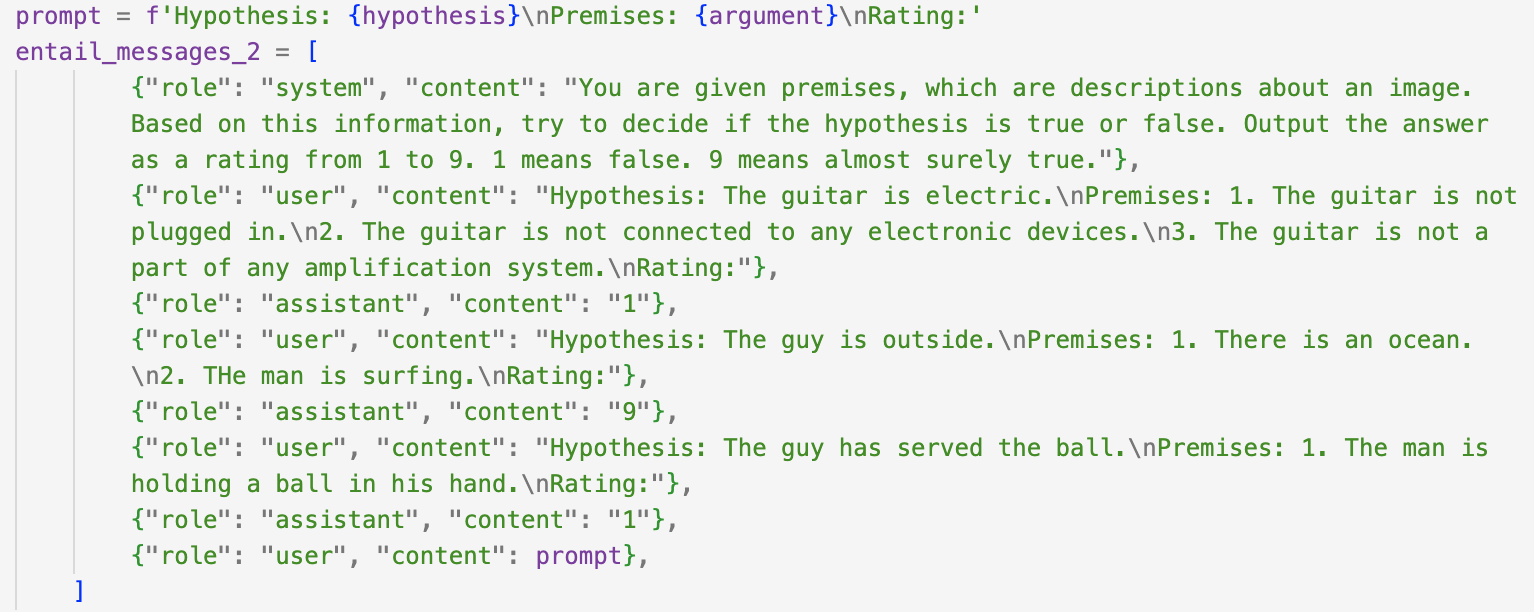}
  \caption{The prompt we use to calculate entailment scores through LLM model as an example for the question in Figure \ref{fig:example}.}
  \label{fig:prompt_entail}
  % \vspace{-em}
\end{figure}

We include the prompt we used to query the NLI model for entailment score as in Figure~\ref{fig:prompt_entail}.

\subsection{Dataset Annotation Details}
\label{sec:app:data}
\begin{figure*}[t]
  \centering
  \includegraphics[width=1\textwidth]{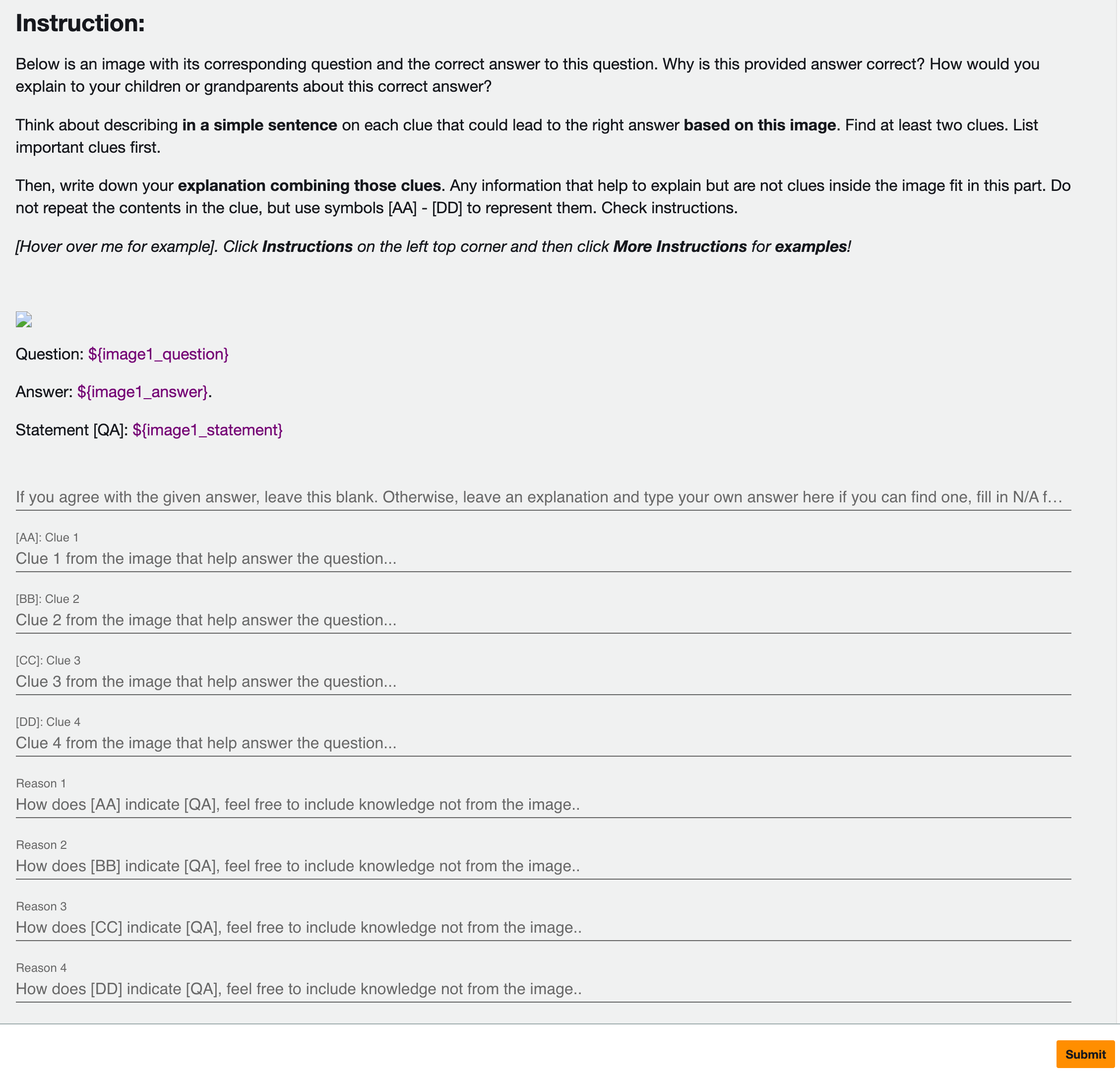}
  \caption{The annotation guidelines we give to mturk workers.}
  \label{fig:mturk}
  \vspace{-0.5em}
\end{figure*}

\begin{figure*}[t]
  \centering
  \includegraphics[width=1\textwidth]{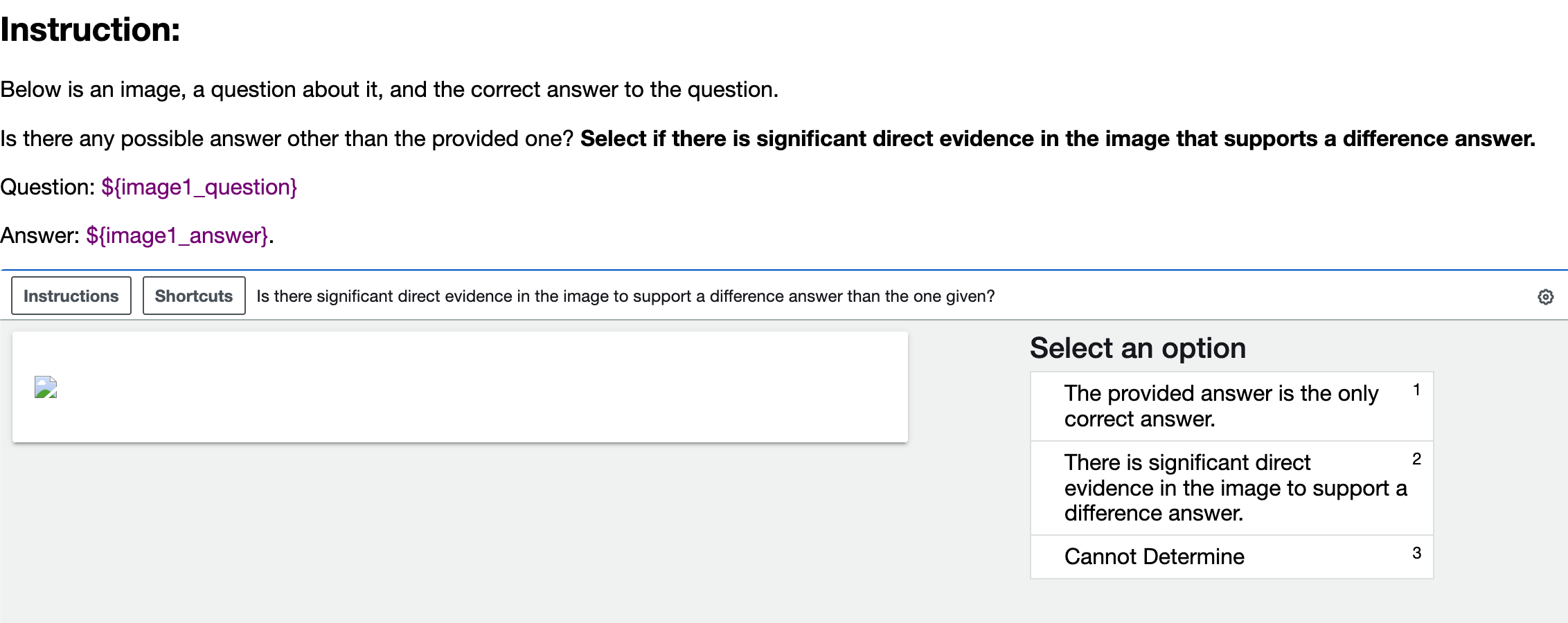}
  \caption{The annotation guidelines we give to mturk workers for filtering out ambiguous ones.}
  \label{fig:mturk_filter}
  \vspace{-0.5em}
\end{figure*}

For every the crowd-source annotation, we offer \$0.7 for each single entry, which is equivalent to \$14 hour pay.
The annotation guidelines are as shown in Figures \ref{fig:mturk} and \ref{fig:mturk_filter}.
Examples of annotated data are shown in Figure \ref{fig:collected_data}. 
% And Figure \ref{fig:prompt_vqa} denotes an example prompt we use to evaluate dataset quality with LLMs. 

% \begin{figure}[h]
%   \centering
%   \includegraphics[width=0.5\textwidth]{figures/prompt_vqa.png}
%   \caption{Example prompt we use to query the LLM model for direct visual question answering, for the example data entry in Figure \ref{fig:data_example}. We hide the 4-shot in-context examples here for space concerns. The answer below the line is LLM output.}
%   \label{fig:prompt_vqa}
%   \vspace{-1em}
% \end{figure}

% \begin{figure*}[t]
%   \centering
%   \includegraphics[width=0.8\textwidth]{figures/data_example.pdf}
%   \caption{An example annotated data entry.}
%   \label{fig:data_example}
%   \vspace{-0.5em}
% \end{figure*}

\subsection{Blip-2 Hyperparameters}
\label{sec:app:hyper}
When fine-tuning on the large weak supervision dataset, we apply a linear warmup of the learning rate during the initial 3K steps, increasing from $1\mathrm{e}{-8}$ to $1\mathrm{e}{-6}$, followed by a cosine decay with a minimum learning rate of $1\mathrm{e}{-8}$.

\begin{table}[t]
\centering
\begin{tabular}{l|c}
\toprule
\multicolumn{2}{c}{Clue Generation Stage 1}\\
\midrule
LLM & FlanT5 $\mathrm{XL}$ \\
Fine-tuning epochs & 5\\
Warmup steps & 3081\\
Learning rate & $5 e-6$\\
Weight decay & 0.01\\
Image resolution & 224\\
Prompt & "Question: \{\} Related Clues:"\\
Inference beam size & 5\\
\bottomrule
\end{tabular}
\begin{tabular}{l|c}
\multicolumn{2}{c}{Clue Generation Stage 2}\\
\midrule
LLM & FlanT5 $\mathrm{XL}$ \\
Fine-tuning epochs & 3\\
Warmup steps & 255\\
Learning rate & $5 e-8$\\
Weight decay & 0.01\\
Image resolution & 224\\
Prompt & "Question: \{\} Related Clues:"\\
Inference beam size & 5\\
\bottomrule
\end{tabular}
\begin{tabular}{l|c}
\multicolumn{2}{c}{Answer Proposal}\\
\midrule
LLM & FlanT5 $\mathrm{XL}$ \\
Fine-tuning epochs & 5\\
Warmup steps & 1000\\
Learning rate & $1 e-5$\\
Weight decay & 0.05\\
Image resolution & 224\\
Prompt & "Question: \{\} Possible Answers:"\\
Inference beam size & 5\\
\bottomrule
% \end{tabular}
% \caption{Hyperparameters for fine-tuning BLIP-2.}
% \label{tab:blip2_hyper}
% \end{table}

% \begin{table}[t]
% \centering
% \begin{tabular}{l|c}
\multicolumn{2}{c}{Baseline VQA}\\
\midrule
LLM & FlanT5 $\mathrm{XL}$ \\
Fine-tuning epochs & 5\\
Warmup steps & 1000\\
Learning rate & $1 e-5$\\
Weight decay & 0.05\\
Image resolution & 224\\
Prompt & "Question: \{\} Answer:"\\
Inference beam size & 5\\
\bottomrule
\end{tabular}
\caption{Hyperparameters for fine-tuning BLIP-2 (continued).}
\label{tab:blip2_hyper}
\end{table}

\end{document}